\ifdefined\XeTeXversion\else\pdfoutput=1 \fi
\documentclass{article}
\usepackage[T1]{fontenc}
\usepackage{times}
\usepackage{arxiv_preprint}

\usepackage{hyperref}
\hypersetup{hidelinks}
\usepackage{url}
\usepackage{booktabs}
\usepackage{amsfonts}
\usepackage{nicefrac}
\usepackage{microtype}
\usepackage{xcolor}
\usepackage{graphicx}
\usepackage{amsmath}
\usepackage{amssymb}
\usepackage{multirow}
\usepackage{amsthm}
\usepackage{tikz}
\usetikzlibrary{shapes.geometric, arrows.meta, positioning, fit, backgrounds}
\usepackage{pgfplots}
\pgfplotsset{compat=1.17}

\graphicspath{{figures/}}

\newtheorem{proposition}{Proposition}
\newtheorem{corollary}{Corollary}

\papertitle{Where Compositionality Hides:\\[2pt]
Shared-Content Interference in Vision-Language Retrieval}

\authorline{%
  Chimata Anudeep\affmark{1}\thanks{Equal contribution.}\hspace{2.2em}
  Harsh Vardhan Poddar\affmark{1}\footnotemark[1]%
}
\affiliationline{%
  \affmark{1}BITS Pilani, K.\,K. Birla Goa Campus%
}

\newcommand{\Cmean}{C_{\text{mean}}}
\newcommand{\Chat}{\hat{C}_{\text{mean}}}

\begin{document}

\maketitle

\begin{abstract}
Paired compositional benchmarks such as Winoground ask each image to prefer its matching caption and each caption to prefer its matching image. A failed Group decision can come from a wrong caption preference or from the cross-image comparison alone, and Group Score merges the two. Choosing between retraining the encoder and changing the scoring rule therefore requires separating them. We study mean-erasure, which projects each frozen image embedding away from the mean of the two normalized candidate captions. We prove that the projection preserves every within-image caption ranking and makes Group Score equal baseline Text Score. On Winoground, mean-erasure raises CLIP ViT-B/32 Group Score from $9\%$ to $31\%$, and five CLIP and SigLIP checkpoints gain $21$ to $27$ percentage points. ColorSwap shows a similar gain for color--object binding. A score decomposition shows that about $60\%$ of Image failures already score the correct joint assignment above the swapped one and fail only on an offset between the two images. Fixed corrections and a learned bilinear scorer gain under one point. A direction estimated from retrieved captions gains about five points with two captions and falls below baseline with sixteen. Before retraining an encoder for paired retrieval, measure the Group loss that mean-erasure recovers: gains beyond baseline Text Score require changing the caption decisions themselves.
\end{abstract}

\section{Introduction}
\label{sec:intro}

\begin{figure}[tbp]
    \centering
    \includegraphics[width=\textwidth,trim=0 20 0 60,clip]{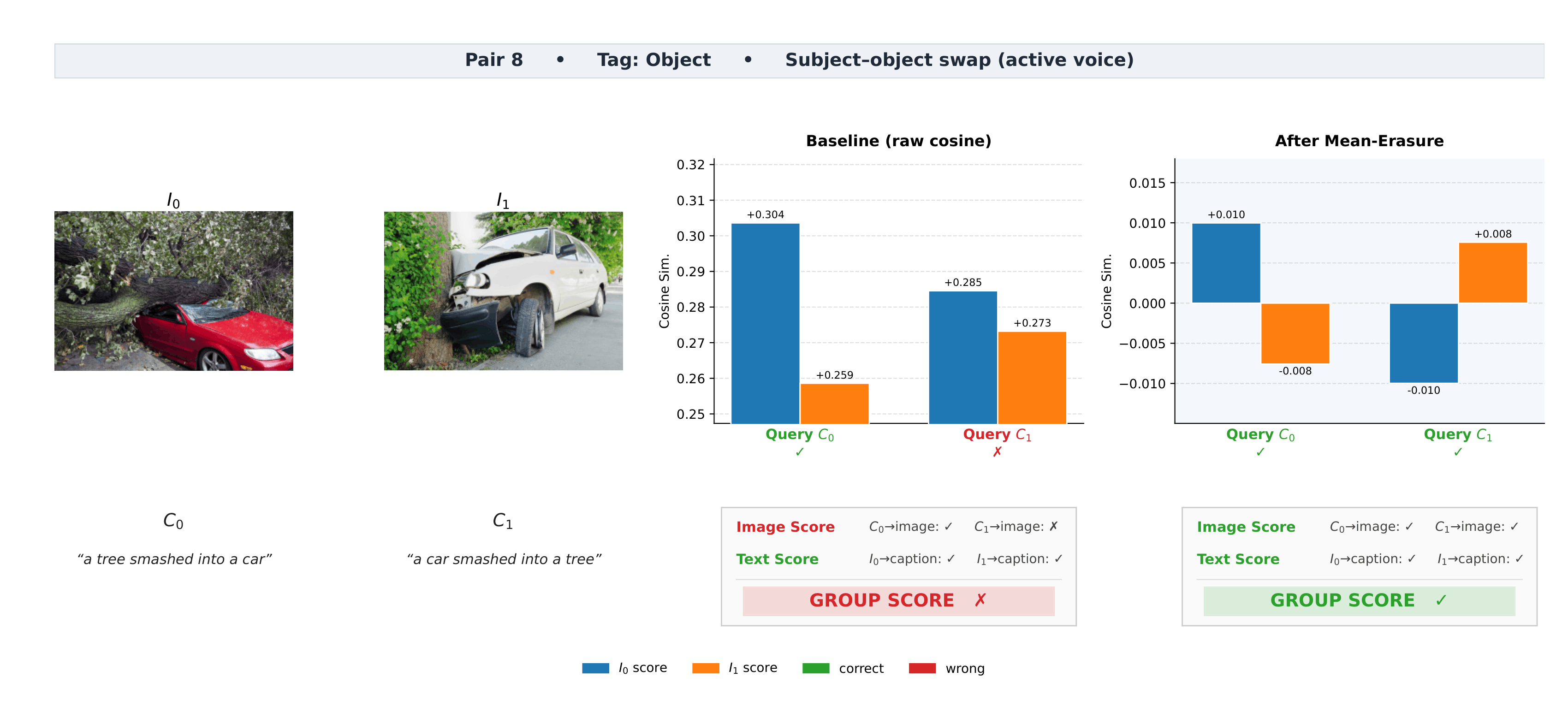}
    \caption{\textbf{A Group failure that mean-erasure recovers (Winoground pair~8).} Both images prefer their correct caption, so Text passes at baseline. Caption $C_1$ prefers the wrong image, so Image and Group fail (middle). After the caption-pair mean is erased from both images, each caption selects its own image and Group passes (right). Blue and orange bars are the scores of $I_0$ and $I_1$.}
    \label{fig:teaser}
\end{figure}

Compositional retrieval requires telling apart images and captions that contain the same objects but assign them different roles or relations. Winoground~\cite{thrush2022winoground} tests this with two images and two captions built from the same words. Its Group Score requires each image to prefer its matching caption and each caption to prefer its matching image. A pair can therefore fail Group even when both images already prefer their correct captions (Figure~\ref{fig:teaser}). Which of the two decisions fails determines which part of a vision--language model needs work.

Contrastive Language-Image Pre-training (CLIP)~\cite{radford2021learning} and related dual encoders~\cite{zhai2023sigmoid,cherti2023reproducible} score pooled image and text vectors by cosine similarity. Training with hard negatives improves compositional retrieval~\cite{yuksekgonul2022when,doveh2023teaching}. Learned alignment and matching procedures also recover performance from frozen representations~\cite{koishigarina2025clipbow,zhu2025matching,miranda2026revisiting}. Their reported Group gain mixes two sources: better caption preferences and better cross-image comparisons. A diagnostic that chooses between them must first identify which decisions fail under the current scoring rule.

\begin{figure}[tbp]
    \centering
    \begin{tikzpicture}[vec/.style={thick, -latex}, scale=0.72]
        \coordinate (O) at (0,0);
        \coordinate (C0) at (3.06, 2.57);
        \coordinate (C1) at (3.06,-2.57);
        \coordinate (Cm) at (3.06, 0);
        \coordinate (I0) at (2.80, 0.60);
        \coordinate (I0p) at (0, 0.60);
        \draw[vec, blue!70!black, line width=1.1pt] (O) -- (Cm)
            node[pos=0.55, below] {\small $\Cmean$};
        \draw[vec, purple!80!black, line width=1.1pt] (Cm) -- (C0)
            node[midway, right] {\small $+\Delta$};
        \draw[vec, purple!80!black, line width=1.1pt] (Cm) -- (C1)
            node[midway, right] {\small $-\Delta$};
        \draw[dashed, black!60, -latex] (O) -- (C0) node[above right] {\small $C_0$};
        \draw[dashed, black!60, -latex] (O) -- (C1) node[below right] {\small $C_1$};
        \draw[vec, densely dashed, teal!80!black, line width=1.1pt] (O) -- (I0)
            node[pos=0.85, below right] {\small $I_0$};
        \draw[vec, orange!90!black, line width=1.4pt] (O) -- (I0p)
            node[above left] {\small $I_0'$};
        \draw[dotted, thick, orange!90!black] (I0) -- (I0p);
    \end{tikzpicture}
    \caption{\textbf{Mean-erasure geometry.} The dotted segment is the component of $I_0$ along $\Cmean$, which the projection removes. Within the caption span, $I'_0$ lies along $\Delta$. In $\mathbb R^d$, the residual can also keep a component orthogonal to both captions.}
    \label{fig:geometry}
\end{figure}
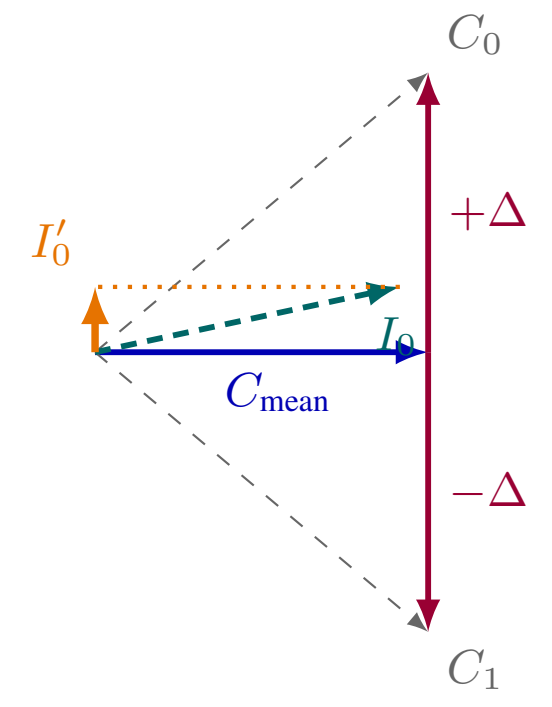

We take the mean of the two candidate-caption embeddings and project each frozen image embedding away from it (Figure~\ref{fig:geometry}). For normalized captions, this mean is orthogonal to the caption difference. Removing it therefore leaves each image's caption preference unchanged while changing the cross-image comparisons. After the projection, every pair that passes baseline Text also passes Group. The gap between baseline Text and baseline Group thus measures the Group loss that a change of scoring rule can recover.

We make four contributions.
\begin{itemize}
\item \textbf{A paired scoring diagnostic.} We prove that mean-erasure preserves caption rankings and makes Group Score equal baseline Text Score, which sets the ceiling for any ranking-preserving correction (Section~\ref{sec:formal}).
\item \textbf{A reusable error decomposition.} We split each Image failure into a joint-matching margin and a cross-image offset. Any dual encoder's score matrix can be tested this way (Section~\ref{sec:offset_results}).
\item \textbf{Evidence for candidate dependence.} We compare the paired direction with an unrelated direction, a noun-prompt direction, five fixed corrections, a learned bilinear scorer, and retrieval-derived directions (Sections~\ref{sec:controls}, \ref{sec:generic}, \ref{sec:learned}, and~\ref{sec:foilfree}).
\item \textbf{Effectiveness across checkpoints and benchmarks.} We measure Group gains of $21$ to $27$ percentage points on five Winoground checkpoints, with similar gains on ColorSwap (Sections~\ref{sec:wino} and~\ref{sec:colorswap}).
\end{itemize}
\noindent\textbf{Reproducibility.} The ancillary files of this submission contain the scoring and audit code, and Appendix~\ref{app:replication_notes} states the evidence behind each comparison.

Before retraining an encoder for paired retrieval, use mean-erasure to separate recoverable cross-image scoring errors from caption-ranking errors. Group gains beyond baseline Text Score require an intervention that changes those caption decisions.

Section~\ref{sec:related} positions the diagnostic against benchmarks, training methods, and readouts. Section~\ref{sec:method} defines mean-erasure and proves its decision boundary. Section~\ref{sec:experiments} measures recovery, decomposes the remaining failures, and compares alternative corrections. Section~\ref{sec:discussion} states the limits and the next intervention.

\section{Related Work}
\label{sec:related}

\paragraph{Compositional benchmarks and their formats.}
Compositional failures of dual encoders are documented by ARO~\cite{yuksekgonul2022when}, VL-CheckList~\cite{zhao2022vlchecklist}, CREPE~\cite{ma2023crepe}, SugarCrepe~\cite{hsieh2023sugarcrepe}, SugarCrepe++~\cite{dumpala2024sugarcrepepp}, and VALSE~\cite{parcalabescu2022valse}. These benchmarks ask one image to choose between two captions ($1\times2$). \citet{hsieh2023sugarcrepe} show that ARO's hard negatives contain distributional artifacts, and \citet{udandarao2025crepe} show that blind heuristics rival CLIP on most benchmarks of this kind. Winoground~\cite{thrush2022winoground} and ColorSwap~\cite{burapacheep2024colorswap} instead pair two images with two captions ($2\times2$), and BiVLC~\cite{miranda2024bivlc} converts SugarCrepe to this format with a generated negative image. The paired format adds cross-image comparisons, and Group Score merges their errors with caption-ranking errors. Mean-erasure separates the two. VisMin~\cite{awal2024vismin} and ConMe~\cite{huang2024conme} offer complementary structural evaluations.

\paragraph{Training-time recovery.}
Fine-tuning with hard negatives~\cite{yuksekgonul2022when,doveh2023teaching}, synthetic negative image--caption pairs~\cite{patel2024tripletclip}, or multi-image objectives~\cite{peleg2025advancing} improves compositional scores. FSC-CLIP~\cite{oh2024fscclip}, CE-CLIP~\cite{zhang2024ceclip}, and READ-CLIP~\cite{kwon2025readclip} add complementary objectives. These methods change the encoder, so a Group gain cannot be attributed to caption decisions or to cross-image comparisons. Mean-erasure keeps the encoder frozen and changes only the cross-image comparison.

\paragraph{Readouts over frozen features.}
Image and text embeddings occupy separate cones~\citep{liang2022mind}, and CLIP composes single objects but degrades when binding is required~\citep{lewis2024clip}. \citet{kang2025clipideal} prove that embeddings built by additive superposition of object parts cannot solve binding, and propose dense per-patch similarity maps. A globally trained linear map recovers the attribute--object binding that CLIP's text encoder encodes linearly~\citep{koishigarina2025clipbow}. \citet{miranda2026revisiting} train a transformer over frozen patch and token features. In the bag-of-concepts account of \citet{alshehri2026similarity}, similarity approximates mean-pooled concept evidence. The proposed readouts need training or patch-level features. Mean-erasure uses only the pooled embeddings and the two candidate captions, and its Group--Text identity states which failures it leaves. \citet{bhattacharya2026inference} parse scene graphs with a multimodal large language model, outside the dual-encoder setting studied here.

\paragraph{Joint matching.}
\citet{zhu2025matching} introduce GroupMatch and Test-Time Matching, which score the correct joint assignment of two images and two captions against the swapped one. Row centering of the cosine score matrix gives the same Image criterion. Mean-erasure also normalizes each projected image, and the two row scales can change Image outcomes (Appendix~\ref{app:chance}). Group still equals baseline Text under both operations, so the diagnostic identifies the remaining caption errors under the original Group criterion.

\paragraph{Fixed scoring corrections.}
Inverted softmax normalizes retrieval scores with a fixed reference-caption bank~\cite{smith2017offline}. Activation steering~\cite{turner2023activation} and text-based decomposition of CLIP image features~\cite{gandelsman2024interpreting} motivate our geometric controls. These corrections apply one transformation to every pair. Mean-erasure derives its direction from the pair being scored, and Section~\ref{sec:generic} measures the difference for the tested corrections.

\section{Mean-Erasure and Its Decision Boundary}
\label{sec:method}

Mean-erasure isolates the Group loss caused by cross-image comparisons and leaves every within-image caption decision unchanged. This section defines the projection, proves that boundary, and states its cost.

\subsection{Setup and Notation}
\label{sec:setup}

Let $I\in\mathbb R^d$ and $C\in\mathbb R^d$ denote $L_2$-normalized image and text embeddings, taken after pooling and the projection head. For normalized vectors, $\mathrm{sim}(I,C)=I^\top C$. All encoder weights stay frozen.

For a candidate caption pair $(C_0,C_1)$, define
\begin{equation}
    \Cmean = \tfrac{1}{2}(C_0 + C_1), \qquad
    \Delta = \tfrac{1}{2}(C_0 - C_1).
    \label{eq:decomp}
\end{equation}
$\Cmean$ is the component shared by the two caption vectors, and $\Delta$ is their contrast. \textit{Mean-erasure} projects an image embedding orthogonally away from $\Cmean$:
\begin{equation}
    I' = I - (I \cdot \Chat)\,\Chat, \qquad
    \Chat = \Cmean / \|\Cmean\|.
    \label{eq:erasure}
\end{equation}
The transformed score is $\mathrm{sim}(I',C)=I'^{\top}C/\|I'\|$, which requires $\Cmean\ne0$ and $\|I'\|>0$.

\paragraph{Evaluation formats.}
A $1\times2$ benchmark presents one image and two captions and asks the model to rank the correct caption higher (SugarCrepe, ARO, VALSE). A $2\times2$ benchmark presents two images and two captions that share content (Winoground, ColorSwap). Write $s_{ij}=\mathrm{sim}(I_i,C_j)$, with $I_0$ matched to $C_0$ and $I_1$ matched to $C_1$. A pair passes \textbf{Text} when $s_{00}>s_{01}$ and $s_{11}>s_{10}$, so each image ranks its caption first. It passes \textbf{Image} when $s_{00}>s_{10}$ and $s_{11}>s_{01}$, so each caption ranks its image first. It passes \textbf{Group} when it passes both. Each Score is the percentage of pairs that pass, and all comparisons are strict. Appendix~\ref{app:notation} collects the notation.

\subsection{Caption Rankings and Cross-Image Comparisons}
\label{sec:formal}

\begin{proposition}[$1\times2$ invariance]
\label{prop:invariance}
Let $C_0, C_1$ be $L_2$-normalized. For any image $I$ with a nonzero projection residual and $\Cmean\ne0$, mean-erasure (Equation~\ref{eq:erasure}) leaves the ranking of $\mathrm{sim}(I, C_0)$ against $\mathrm{sim}(I, C_1)$ unchanged.
\end{proposition}

\begin{proof}
Equal caption norms give
\[
\Cmean\cdot\Delta=\tfrac14(\|C_0\|^2-\|C_1\|^2)=0.
\]
The projection in Equation~\ref{eq:erasure} therefore satisfies
\[
I'\cdot\Delta=I\cdot\Delta,\qquad
\mathrm{sim}(I',C_0)-\mathrm{sim}(I',C_1)
=\frac{2I\cdot\Delta}{\|I'\|}.
\]
The positive denominator preserves the sign of the original caption-score difference.
\end{proof}

\begin{corollary}[Cross-image offset]
\label{cor:noninvariance}
Under the assumptions of Proposition~\ref{prop:invariance}, the unnormalized cross-image difference is
\[
s_{00}-s_{10}=(I_0-I_1)\cdot\Cmean+(I_0-I_1)\cdot\Delta.
\]
Mean-erasure removes the first term. Normalization then divides the two image scores by their residual norms $\|I'_0\|$ and $\|I'_1\|$.
\end{corollary}

The first term is an offset that favors whichever image lies closer to the shared caption content, regardless of the caption contrast. Corollary~\ref{cor:noninvariance} shows that this offset enters every cross-image comparison and none of the within-image ones.

\begin{corollary}[Group--Text identity]
\label{cor:groupscore}
For normalized captions, a nonzero caption mean, and nonzero image residuals, a Text pass after erasure gives
\[
I'_0\cdot\Delta>0>I'_1\cdot\Delta.
\]
Both Image comparisons reduce to
\[
\frac{I'_0\cdot\Delta}{\|I'_0\|}>
\frac{I'_1\cdot\Delta}{\|I'_1\|},
\]
which follows from these signs. Group is the conjunction of Text and Image. With Proposition~\ref{prop:invariance},
\begin{equation}
\mathrm{Group}_{\rm ME}=\mathrm{Text}_{\rm ME}=\mathrm{Text}_{\rm base}.
\label{eq:group_text}
\end{equation}
\end{corollary}

\paragraph{Assumptions.}
Normalized captions hold by construction for CLIP and SigLIP embeddings. A zero caption mean requires $C_1=-C_0$, and a zero residual requires $I$ to be parallel to $\Chat$. Appendix~\ref{app:replication_notes} gives the audit code's handling of both degenerate cases.

\subsection{Implications and Cost}
\label{sec:implications}

\textbf{Baseline Text Score is the ceiling.} Every baseline Group pass is a Text pass, so mean-erasure recovers the Text-pass pairs that fail Group. Pairs with a wrong caption preference stay Group failures. Stronger erasure cannot lift this ceiling while caption rankings are preserved. A method that targets higher Group Score must change those rankings, through a different readout, better cross-modal alignment, or new representations. The identity is specific to the $2\times2$ format, and Appendix~\ref{app:larger} gives a $3\times3$ counterexample.

\paragraph{A complementary diagnostic.}
Exchanging the roles of images and captions gives a second diagnostic. Define $I_{\rm mean}=\frac12(I_0+I_1)$ and project each caption away from it:
\begin{equation}
    C' = C - (C \cdot \hat{I}_{\rm mean})\,\hat{I}_{\rm mean},
    \qquad \hat{I}_{\rm mean} = I_{\rm mean} / \|I_{\rm mean}\|.
    \label{eq:image_erasure}
\end{equation}
For normalized images, a nonzero image mean, and nonzero caption residuals, Proposition~\ref{prop:invariance} and Corollary~\ref{cor:groupscore} with the roles exchanged give $\mathrm{Image}_{\rm C\text{-}erased}=\mathrm{Group}_{\rm C\text{-}erased}=\mathrm{Image}_{\rm base}$. The two operations therefore recover different pass sets: baseline Text passes and baseline Image passes. Table~\ref{tab:overlap} gives their overlap for CLIP ViT-B/32. Caption-mean erasure recovers $88$ pairs that image-mean erasure misses, and image-mean erasure recovers $9$ that caption-mean erasure misses. Two thirds of the pairs pass neither, so they fail both a caption decision and an image decision.

\begin{table}[tbp]
\centering
\caption{Group-pass sets of the two erasure operations for CLIP ViT-B/32 on Winoground ($n=400$). Counts follow from the baseline totals (Table~\ref{tab:audit_wino}) and the two identities.}
\label{tab:overlap}
\begin{tabular}{lcc}
\toprule
\textbf{Group-pass set} & $\mathbf{n}$ & \textbf{\%} \\
\midrule
Baseline & 36 & 9.00 \\
Caption-mean erasure of images & 124 & 31.00 \\
Image-mean erasure of captions & 45 & 11.25 \\
\midrule
Both operations & 36 & 9.00 \\
Only caption-mean erasure & 88 & 22.00 \\
Only image-mean erasure & 9 & 2.25 \\
Neither operation & 267 & 66.75 \\
\bottomrule
\end{tabular}
\end{table}

\paragraph{Implementation and cost.}
For a pair, average its two normalized captions, normalize the mean, project each image with Equation~\ref{eq:erasure}, and normalize each residual before cosine scoring. On cached embeddings of dimension $d$, the projection adds $O(d)$ arithmetic and $O(d)$ temporary storage per pair, and no encoder forward passes.

\section{Experiments}
\label{sec:experiments}

The experiments first measure how much Group loss paired scoring recovers and why the remaining Image failures occur. They then test whether the direction must come from the candidate pair, how learned and generative scorers compare, and how far a retrieved direction can replace the supplied pair.

\subsection{Evaluation Setup}
\label{sec:eval_setup}

Winoground contains $400$ image--caption pairs. We evaluate OpenAI CLIP ViT-B/32 and ViT-L/14, LAION CLIP ViT-B/32, SigLIP-base, and SigLIP-SO400M (identifiers in Appendix~\ref{app:replication_notes}). ColorSwap tests color--object binding on $300$ groups. Each baseline and its mean-erasure variant use the same frozen embeddings. Scores are percentages of pairs that pass the strict criteria in Section~\ref{sec:setup}.

The five-checkpoint Winoground scores and the direction control come from a saved aggregate output, from which we reconstruct integer pass counts and confidence intervals. Appendix~\ref{app:replication_notes} states the source of every other table.

\subsection{Paired Recovery on Winoground}
\label{sec:wino}

\textbf{Mean-erasure recovers $21$ to $27$ percentage points of Group Score on all five checkpoints.} On CLIP ViT-B/32, Group Score rises from $9.0\%$ to $31.0\%$ and Text Score stays at $31.0\%$ (Table~\ref{tab:erasure}). The recovered counts follow from the baseline pass totals and the Group--Text identity. The Wilson intervals describe sampling variation across benchmark pairs. For CLIP ViT-B/32, the $69\%$ of pairs that still fail Group are the baseline Text failures. Figure~\ref{fig:teaser} shows one recovered pair, and Appendix~\ref{app:qual_success} shows two more.

\textbf{Recovery spans relation and object swaps.} Object swaps gain $27.7$ percentage points and relation swaps gain $16.7$ (Table~\ref{tab:category}). The both-swap subset gains $38.5$ points, but its $26$ pairs support only a coarse estimate.

\begin{figure}[tbp]
    \centering
    \includegraphics[width=0.92\textwidth,trim=0 20 0 60,clip]{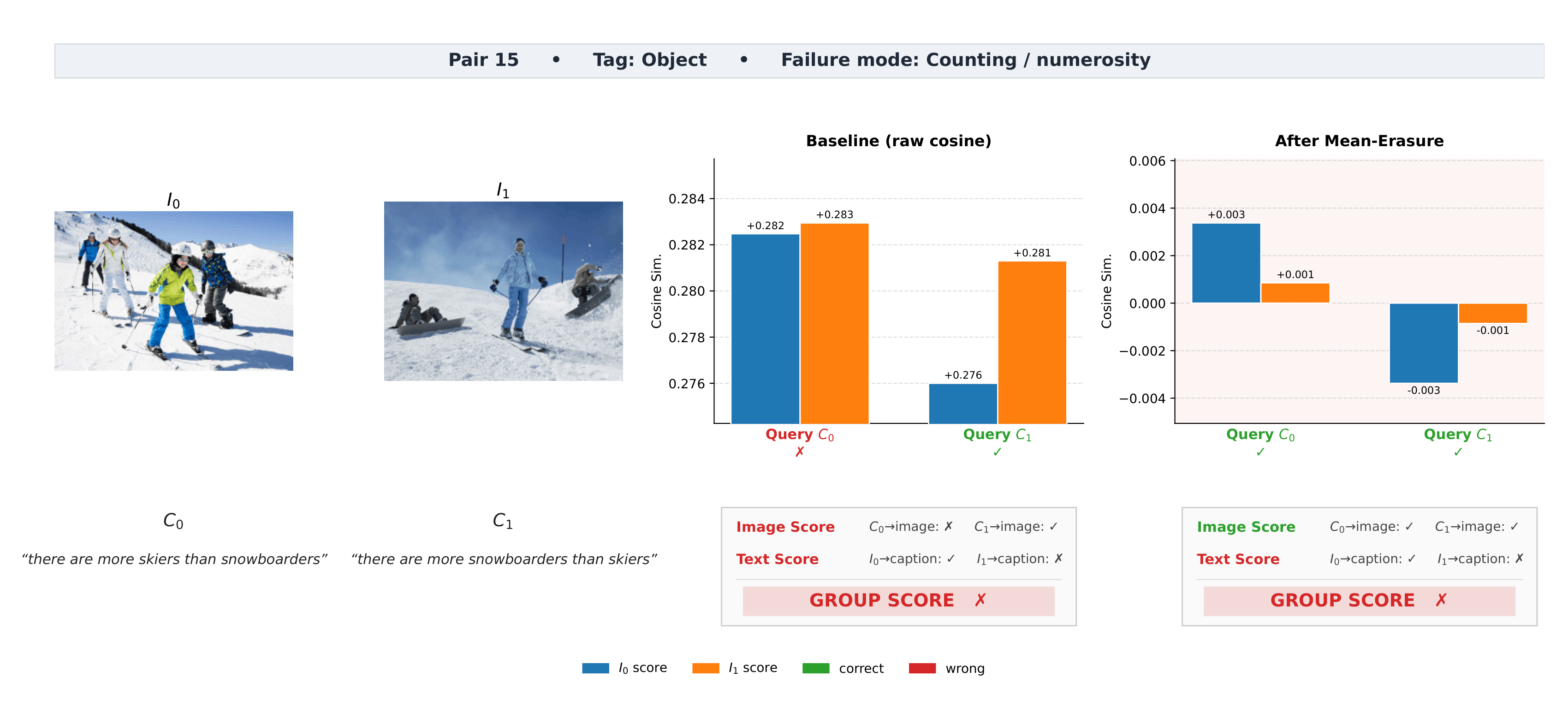}\\[2pt]
    \includegraphics[width=0.92\textwidth,trim=0 20 0 60,clip]{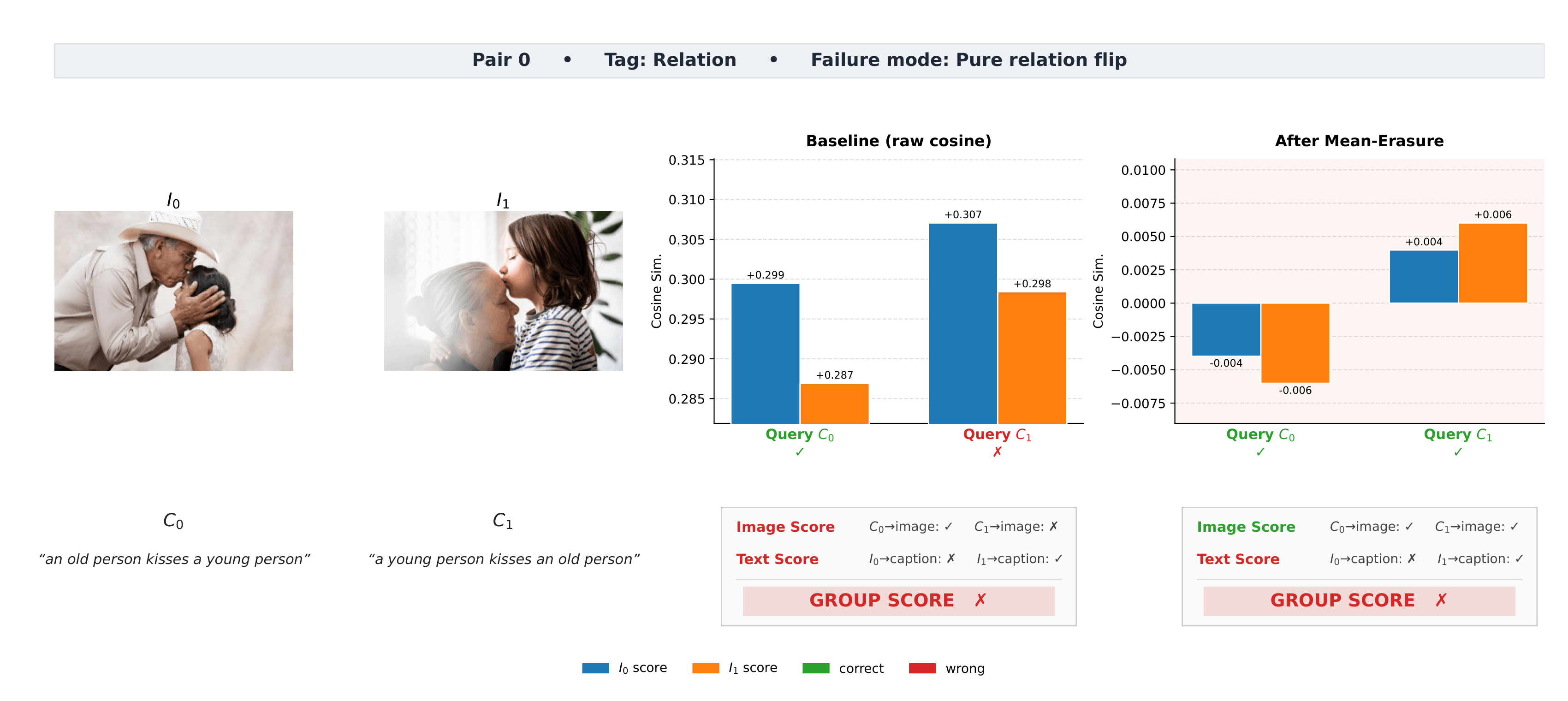}
    \caption{\textbf{Group failures that mean-erasure cannot recover.} Top: Winoground pair~15 (counting). Bottom: pair~0 (relation flip). In both pairs one image prefers the wrong caption, so Text fails, and the projection leaves that preference unchanged (Proposition~\ref{prop:invariance}). Panels follow Figure~\ref{fig:teaser}.}
    \label{fig:qual_fail}
\end{figure}

\textbf{The pairs that stay Group failures have a wrong caption preference.} In Figure~\ref{fig:qual_fail}, pair~15 asks the model to compare counts of skiers and snowboarders. Image $I_1$ prefers the wrong caption, so Text fails at baseline and, by Proposition~\ref{prop:invariance}, after erasure as well, although Image passes after erasure. Pair~0 reverses which person kisses which. Its caption preference also fails, so its Group decision cannot be recovered. Appendix~\ref{app:qual_failures} shows a third failure, a multi-object spatial re-binding.

\begin{table}[tbp]
\centering
\caption{Winoground Group Score (\%), $n=400$. Mean-erasure equals baseline Text. Gain is in percentage points, with a $95\%$ Wilson confidence interval (CI) for the recovered-pair proportion.}
\label{tab:erasure}
\begin{tabular}{lccc}
\toprule
\textbf{Checkpoint} & \textbf{Baseline} & \textbf{Mean-erasure} & \textbf{Gain [95\% CI]} \\
\midrule
CLIP ViT-B/32 (OpenAI) & 9.00 & 31.00 & 22.00 [18.22, 26.32] \\
CLIP ViT-B/32 (LAION) & 8.25 & 35.00 & 26.75 [22.65, 31.29] \\
CLIP ViT-L/14 (OpenAI) & 7.25 & 28.25 & 21.00 [17.29, 25.26] \\
SigLIP-base & 10.25 & 33.00 & 22.75 [18.91, 27.11] \\
SigLIP-SO400M & 12.25 & 37.25 & 25.00 [21.01, 29.47] \\
\bottomrule
\end{tabular}
\end{table}

\begin{table}[tbp]
\centering
\caption{Winoground Group Score (\%) by swap tag, before and after mean-erasure (CLIP ViT-B/32).}
\label{tab:category}
\begin{tabular}{lccc}
\toprule
\textbf{Swap tag} & $\mathbf{n}$ & \textbf{Baseline} & \textbf{Mean-erasure} \\
\midrule
Relation & 233 & 6.0 & 22.7 \\
Object & 141 & 7.8 & 35.5 \\
Both & 26 & 42.3 & 80.8 \\
\midrule
All & 400 & 9.0 & 31.0 \\
\bottomrule
\end{tabular}
\end{table}

\subsection{Cross-Image Offsets and Joint Matching}
\label{sec:offset_results}

\textbf{Most Image failures already score the correct joint assignment higher.} Write $a_i=(s_{i0}+s_{i1})/2$ for the score component common to both captions and $b_i=(s_{i0}-s_{i1})/2$ for the signed caption contrast. The joint-matching half-margin is $m=b_0-b_1$, and the cross-image offset is $o=|a_0-a_1|$. Baseline Image passes only when $m>o$, while row-centered Image needs only $m>0$ (Appendix~\ref{app:score_audit}). For CLIP ViT-B/32, $213$ of the $355$ baseline Image failures ($60\%$) have $0<m\le o$: the correct assignment wins, but the offset is larger. The share ranges from $60\%$ to $64\%$ across the five checkpoints (Table~\ref{tab:audit_wino}).

\begin{table}[tbp]
\centering
\caption{Winoground score audit (\%), $n=400$. $T,V,G$: baseline Text, Image, and Group. $J$: row-centered Image (two-way joint matching). $V_{\rm ME}$: Image after normalized mean-erasure. $F$: share of baseline Image failures with $0<m\le o$.}
\label{tab:audit_wino}
\begin{tabular}{lcccccc}
\toprule
\textbf{Checkpoint} & $\mathbf{T}$ & $\mathbf{V}$ & $\mathbf{G}$ & $\mathbf{J}$ & $\mathbf{V}_{\rm ME}$ & $\mathbf{F}$ \\
\midrule
CLIP ViT-B/32 (OpenAI) & 31.00 & 11.25 & 9.00 & 64.50 & 64.50 & 60.0 \\
CLIP ViT-B/32 (LAION) & 35.00 & 11.25 & 8.25 & 66.25 & 66.75 & 62.0 \\
CLIP ViT-L/14 (OpenAI) & 28.25 & 10.50 & 7.25 & 65.50 & 65.50 & 61.5 \\
SigLIP-base & 33.00 & 13.00 & 10.25 & 67.25 & 67.00 & 62.4 \\
SigLIP-SO400M & 37.25 & 16.50 & 12.25 & 70.00 & 70.25 & 64.1 \\
\bottomrule
\end{tabular}
\end{table} Joint matching and the original Image criterion therefore test different properties of the same score matrix, and a failure analysis should report both. Normalized mean-erasure can also differ from row-centered Image, because the residual norms rescale the two rows. The $J$ and $V_{\rm ME}$ columns differ by at most $0.5$ points.

\subsection{Candidate Direction and Erasure Strength}
\label{sec:controls}

\textbf{The direction must come from the candidate pair.} Erasing the mean of an unrelated caption pair lowers CLIP ViT-B/32 Group Score to $7.0\%$, below the $9.0\%$ baseline, and lowers Text Score to $24.0\%$ (Table~\ref{tab:direction_controls}). Erasing a prompt built from the nouns of the caption, which names the shared objects without the pair's second caption, gives a Group Score of $7.25\%$ and a Text Score of $25.5\%$ (Appendix~\ref{app:replication_notes}). The matched direction preserves Text by construction. Recovering Group loss without changing caption decisions thus requires the actual candidate pair.

\begin{table}[tbp]
\centering
\caption{CLIP ViT-B/32 Winoground direction control (\%). The unrelated caption-pair mean and the noun-prompt direction are erased while the original captions are scored. The best scores are in bold.}
\label{tab:direction_controls}
\begin{tabular}{lccc}
\toprule
\textbf{Erased direction} & \textbf{Text} & \textbf{Image} & \textbf{Group} \\
\midrule
None (baseline) & 31.0 & 11.25 & 9.0 \\
Matched caption-pair mean & 31.0 & \textbf{64.5} & \textbf{31.0} \\
Unrelated caption-pair mean & 24.0 & 10.5 & 7.0 \\
Noun prompt & 25.5 & 10.25 & 7.25 \\
\bottomrule
\end{tabular}
\end{table}

\textbf{Full erasure reaches the ceiling in the strength sweep.} With partial strength $\alpha\in\{0,0.1,\ldots,1\}$, Group Score rises with $\alpha$ and reaches baseline Text at $\alpha=1$ (Figure~\ref{fig:sweep}; values in Appendix~\ref{app:sensitivity}). Image Score rises fastest above $\alpha=0.8$, from $30.8\%$ to $64.5\%$.

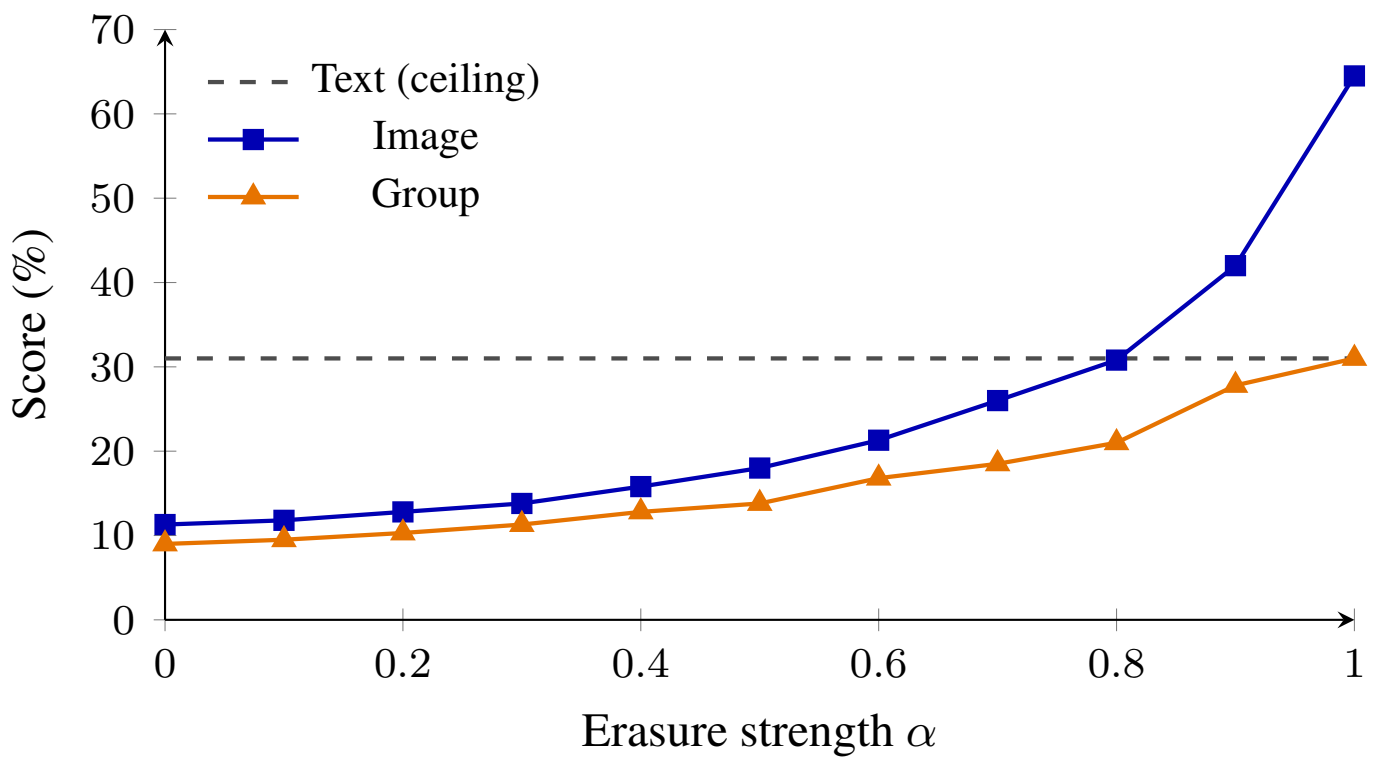
\begin{figure}[tbp]
\centering
\begin{tikzpicture}
\begin{axis}[
    width=0.62\textwidth, height=0.36\textwidth,
    xlabel={Erasure strength $\alpha$}, ylabel={Score (\%)},
    xmin=0, xmax=1, ymin=0, ymax=70,
    xtick={0,0.2,0.4,0.6,0.8,1.0}, ytick={0,10,20,30,40,50,60,70},
    legend style={at={(0.02,0.98)}, anchor=north west, font=\footnotesize, draw=none, fill=none},
    tick label style={font=\footnotesize}, label style={font=\small},
    axis lines=left, grid=none,
]
\addplot[black!70, dashed, thick, mark=none] coordinates {(0,31.0) (1,31.0)};
\addlegendentry{Text (ceiling)}
\addplot[blue!70!black, thick, mark=square*, mark size=1.6pt] coordinates {
(0,11.3) (0.1,11.8) (0.2,12.8) (0.3,13.8) (0.4,15.8) (0.5,18.0) (0.6,21.3) (0.7,26.0) (0.8,30.8) (0.9,42.0) (1.0,64.5)};
\addlegendentry{Image}
\addplot[orange!90!black, thick, mark=triangle*, mark size=2pt] coordinates {
(0,9.0) (0.1,9.5) (0.2,10.3) (0.3,11.3) (0.4,12.8) (0.5,13.8) (0.6,16.8) (0.7,18.5) (0.8,21.0) (0.9,27.8) (1.0,31.0)};
\addlegendentry{Group}
\end{axis}
\end{tikzpicture}
\caption{\textbf{Partial erasure on Winoground (CLIP ViT-B/32).} Group Score rises with the erasure strength $\alpha$ and meets the Text ceiling (dashed) at full erasure. Text stays constant by Proposition~\ref{prop:invariance}.}
\label{fig:sweep}
\end{figure} Full erasure is therefore the default. The rise is empirical for normalized scores, and Appendix~\ref{app:chance} proves it for unnormalized scores.

\subsection{ColorSwap}
\label{sec:colorswap}

\textbf{The diagnostic recovers a similar Group loss on color--object binding.} The Group gain ranges from about $24$ to $33$ percentage points across the five checkpoints, and every lower confidence bound exceeds $19$ points (Table~\ref{tab:colorswap}). ColorSwap contrasts colors, while Winoground contrasts roles and relations. The paired format still imposes the Group--Text ceiling, so the recovered decisions keep the original caption preferences.

\begin{table}[tbp]
\centering
\caption{ColorSwap Group Score (\%), $n=300$. Gain is in percentage points, with a $95\%$ Wilson CI; both use pass counts reconstructed from the displayed scores.}
\label{tab:colorswap}
\begin{tabular}{lccc}
\toprule
\textbf{Checkpoint} & \textbf{Baseline} & \textbf{Mean-erasure} & \textbf{Gain [95\% CI]} \\
\midrule
CLIP ViT-B/32 (OpenAI) & 12.0 & 35.7 & 23.7 [19.2, 28.8] \\
CLIP ViT-L/14 (OpenAI) & 7.3 & 32.7 & 25.3 [20.7, 30.5] \\
CLIP ViT-B/32 (LAION) & 23.3 & 56.3 & 33.0 [27.9, 38.5] \\
SigLIP-base & 30.3 & 61.3 & 31.0 [26.0, 36.4] \\
SigLIP-SO400M & 37.0 & 67.3 & 30.3 [25.4, 35.8] \\
\bottomrule
\end{tabular}
\end{table}

\subsection{Fixed Scoring Corrections}
\label{sec:generic}

\textbf{The tested fixed corrections gain at most $0.75$ points of Group Score.} The best fixed correction, inverted softmax, scores $9.75\%$ against the $9.0\%$ baseline, with a bootstrap tail of $0.31$ that leaves this small gain unresolved (Table~\ref{tab:interventions}). The fixed geometric transformations score at or below their baselines. At these settings, adapting the direction to the pair outperforms every fixed correction.

Singular value decomposition (SVD) steering scales the coordinates of the visual projection with the smallest singular values. A contrastive activation vector (CAV) adds a direction estimated from $100$ pairs and is evaluated on the other $300$. Mahalanobis reweighting rescales both modalities in the output basis of the visual projection. Two random controls apply the matching geometric operations. Inverted softmax divides by an image-specific sum over a fixed $4{,}126$-caption bank. Appendix~\ref{app:interventions} gives every transformation and parameter.

A bootstrap tail is the fraction of $10{,}000$ paired resamples whose mean Group difference is at or below zero. The tails describe the bootstrap distribution; they are not calibrated test probabilities.

\begin{table}[tbp]
\centering
\caption{CLIP ViT-B/32 Winoground scores (\%) under fixed corrections, with the best in bold. The last three rows use the $300$-pair CAV evaluation split. Paired mean-erasure is the candidate-dependent reference.}
\label{tab:interventions}
\begin{tabular}{lcccc}
\toprule
\textbf{Correction} & \textbf{Text} & \textbf{Image} & \textbf{Group} & \textbf{Bootstrap tail} \\
\midrule
Baseline ($n=400$) & 31.0 & 11.3 & 9.0 & n/a \\
Mean-erasure (paired) & 31.0 & \textbf{64.5} & \textbf{31.0} & $<10^{-4}$ \\
Inverted softmax ($\beta=20$) & 31.0 & 12.5 & 9.75 & 0.31 \\
SVD bottom $20\%$ ($\alpha=2$) & 28.5 & 10.3 & 7.5 & 0.912 \\
Random subspace & 22.3 & 7.8 & 5.8 & 0.981 \\
Mahalanobis reweighting & 20.8 & 2.5 & 1.8 & 1.000 \\
\midrule
Baseline ($n=300$) & 32.0 & 12.3 & 9.3 & n/a \\
CAV ($\alpha=5$) & 33.3 & 11.3 & 9.0 & 0.691 \\
Random pre-projection vector & 31.0 & 12.3 & 9.3 & 0.604 \\
\bottomrule
\end{tabular}
\end{table}

\subsection{Learned and Generative Scorers}
\label{sec:learned}

\textbf{A learned bilinear scorer transfers at most $0.75$ points of Group Score.} The bilinear scorer replaces cosine with $s(I,C)=I^\top MC$, where $M\in\mathbb{R}^{d\times d}$ starts at the identity. We train $M$ on ColorSwap and evaluate it zero-shot on Winoground. Across the weight-decay grid, the Winoground Group gain ranges from $0.25$ to $0.75$ percentage points, and weaker regularization lowers the transfer (Appendix~\ref{app:bilinear}). A candidate-independent scorer trained on one paired benchmark therefore recovers far less than the $22$ points that the paired direction recovers.

\textbf{A generative scorer changes the caption decisions and reaches a similar Group Score.} LLaVA-1.5-7B~\citep{liu2024improved} reads dense CLIP ViT-L/14 features and scores each image--caption combination by its yes-or-no likelihood (Appendix~\ref{app:llava}). Its Winoground Text Score is $46.75\%$, its Image Score is $42.50\%$, and its Group Score is $30.0\%$. The Text Score exceeds the $28.25\%$ of pooled CLIP ViT-L/14, so the generative readout changes caption decisions, which Corollary~\ref{cor:groupscore} requires for gains beyond the ceiling. Its Group Score is within $2$ points of the $28.25\%$ that mean-erasure reaches on the pooled embeddings of the same encoder (Table~\ref{tab:erasure}).

\subsection{Pooled-Feature Probes on SugarCrepe}
\label{sec:probe}

\textbf{A probe on pooled features stays at chance on swap captions.} We train a multilayer perceptron (MLP) probe on concatenated pooled image and caption embeddings to separate matching from mismatched captions on SugarCrepe. On \texttt{swap\_obj}, \texttt{replace\_att}, and \texttt{swap\_att}, held-out accuracy ranges from $50.0\%$ to $51.5\%$, against a chance rate of $50\%$ (Table~\ref{tab:probe} in Appendix~\ref{app:probe}). The swap categories pose the same contrast as Winoground: identical words in a different arrangement. Under this probe, pooled embeddings do not separate swapped captions, so the probe offers no route past the caption-decision ceiling. On the two addition categories, preferring the shorter caption scores above $98\%$ and exceeds the probe, which agrees with the length artifacts reported by \citet{hsieh2023sugarcrepe} and \citet{udandarao2025crepe}.

\subsection{Retrieval-Derived Directions}
\label{sec:foilfree}

\textbf{A retrieved direction gives a smaller gain, which reverses as $K$ grows.} Group Score rises from $9.0\%$ to $14.5\%$ at $K=2$ and falls below baseline at $K=16$ and $K=32$ (Table~\ref{tab:topk}). Text Score drops by $2.5$ to $8$ percentage points across the sweep. Replacing the supplied pair with retrieved captions therefore gives up the caption-ranking guarantee, and more retrieved captions make the approximation worse.

For query image $I$, let $\mathrm{TopK}(I)$ hold its top-$K$ captions under baseline cosine similarity in a fixed gallery. We compute $\mu_K(I)=K^{-1}\sum_{C\in\mathrm{TopK}(I)}C$ and use its normalized direction in Equation~\ref{eq:erasure}, so each query has its own direction. Appendix~\ref{app:topk} specifies the gallery in the accompanying code.

\begin{table}[tbp]
\centering
\caption{Top-$K$ erasure on Winoground (CLIP ViT-B/32). Group Score is in percent; change is in percentage points.}
\label{tab:topk}
\begin{tabular}{lccc}
\toprule
$K$ & \textbf{Group} & \textbf{Change} & \textbf{Bootstrap tail} \\
\midrule
Baseline & 9.00 & n/a & n/a \\
2 & 14.50 & $+5.50$ & 0.0012 \\
4 & 10.50 & $+1.50$ & 0.182 \\
8 & 9.25 & $+0.25$ & 0.469 \\
16 & 8.75 & $-0.25$ & 0.611 \\
32 & 8.25 & $-0.75$ & 0.749 \\
\bottomrule
\end{tabular}
\end{table}

\paragraph{Practical guidance.}
With both candidate captions available, use full mean-erasure ($\alpha=1$) to measure the recoverable Group loss. CLIP ViT-B/32 recovers $22$ percentage points (Table~\ref{tab:erasure}). Baseline Text Score is the ceiling, and further Group gains require fixing failed caption decisions (Corollary~\ref{cor:groupscore}). An MLP probe on pooled features does not fix them (Section~\ref{sec:probe}). Expect no change on $1\times2$ benchmarks (Proposition~\ref{prop:invariance}). On cached embeddings, the projection costs $O(d)$ arithmetic and storage per pair (Section~\ref{sec:implications}). For gallery retrieval, validate $K$ and monitor caption accuracy: the sweep gains $5.5$ points at $K=2$ and loses Group Score at $K=16$ and $K=32$ (Table~\ref{tab:topk}).

\section{Limitations and Extensions}
\label{sec:discussion}

Mean-erasure cannot repair a wrong within-image caption preference (Proposition~\ref{prop:invariance}). The counting and relation-flip pairs in Figure~\ref{fig:qual_fail} stay Group failures for that reason, and $69\%$ of CLIP ViT-B/32 pairs fail Group after erasure. This negative result makes a caption-ranking readout the next target for paired retrieval. The readout needs more than pooled embeddings: an MLP probe on them stays at chance on swap captions (Section~\ref{sec:probe}), while LLaVA's dense-feature readout raises Text Score to $46.75\%$ (Section~\ref{sec:learned}). Extending the correction to an open gallery also requires a reliable estimate of the candidate contrast. The top-$K$ losses in Table~\ref{tab:topk} rule out retrieving more captions as a default remedy. A concrete extension is to estimate directions from validated caption clusters, with pairs held out from cluster construction, and to report caption-ranking preservation alongside gallery retrieval.

\section{Conclusion}
\label{sec:conclusion}

Paired compositional evaluation should separate each image's caption decision from the cross-image scoring rule before it motivates encoder retraining. Mean-erasure gives that separation on frozen embeddings and names the remaining work: Group Score beyond baseline Text requires changing caption preferences, through a readout or representation that improves within-image discrimination. Reporting Text Score next to Group Score therefore shows which of the two a new method improves. A Group gain at unchanged caption decisions is a scoring gain, and only better caption decisions raise the ceiling. For gallery retrieval, estimating the candidate direction becomes its own learning problem, and its evaluation must count caption-ranking losses as well as Group gains.

\section*{Acknowledgments}
We thank Prof.\ Sravan Danda for his detailed feedback on the writing and presentation of this paper.

\bibliography{main}
\bibliographystyle{plainnat}

\clearpage
\appendix

\section{Sensitivity to Erasure Strength}
\label{app:sensitivity}

We sweep a partial-erasure strength $\alpha \in [0,1]$ with
$I'_{\alpha} = I - \alpha (I \cdot \Chat)\Chat$. We evaluate $\alpha\in\{0,0.1,\ldots,1\}$ (Table~\ref{tab:sensitivity}; plotted in Figure~\ref{fig:sweep}). Appendix~\ref{app:chance} derives the unnormalized monotonicity condition. The sweep supports $\alpha=1$ as the full-erasure diagnostic, with no further Group recovery available from preserving the same caption decisions.

\begin{table}[tbp]
\centering
\caption{Winoground scores versus partial-erasure strength $\alpha$
(\texttt{clip-vit-base-patch32}). $\alpha=0$ is baseline; $\alpha=1$ is
full mean-erasure.}
\label{tab:sensitivity}
\begin{tabular}{lccc}
\toprule
$\alpha$ & \textbf{Text (\%)} & \textbf{Image (\%)} & \textbf{Group (\%)} \\
\midrule
0.0 & 31.0 & 11.3 & 9.0 \\
0.1 & 31.0 & 11.8 & 9.5 \\
0.2 & 31.0 & 12.8 & 10.3 \\
0.3 & 31.0 & 13.8 & 11.3 \\
0.4 & 31.0 & 15.8 & 12.8 \\
0.5 & 31.0 & 18.0 & 13.8 \\
0.6 & 31.0 & 21.3 & 16.8 \\
0.7 & 31.0 & 26.0 & 18.5 \\
0.8 & 31.0 & 30.8 & 21.0 \\
0.9 & 31.0 & 42.0 & 27.8 \\
1.0 & 31.0 & 64.5 & 31.0 \\
\bottomrule
\end{tabular}
\end{table}

\section{Content-Agnostic Intervention Details}
\label{app:interventions}

\paragraph{Projection convention.}
Let $h\in\mathbb R^p$ be a pooled image feature before projection. Let $W\in\mathbb R^{d\times p}$ be the visual projection matrix, with projected feature $z=Wh$. Write its thin SVD as $W=U\Sigma V^\top$, with singular values $\sigma_1\ge\cdots\ge\sigma_d$. The columns of $U\in\mathbb R^{d\times d}$ define the output basis. The convention applies to CLIP ViT-B/32, where $p=768$ and $d=512$.

\paragraph{SVD steering \citep{gandelsman2024interpreting}.}
Express each projected image feature in the output basis as $U^\top z$. Multiply the $102$ coordinates with the smallest singular values by $1+\alpha$, using $\alpha=2$. Transform back with $U$. Normalize before cosine scoring. Caption embeddings remain unchanged.

\paragraph{Contrastive activation vectors \citep{turner2023activation}.}
The supplied implementation uses pre-projection features to construct
\[
d_{\rm CAV}=\frac1{100}\sum_{k=1}^{100}(h_0^{(k)}-h_1^{(k)}).
\]
The $100$ construction pairs are selected with seed $42$. Evaluation uses the other $300$ pairs. Normalize $d_{\rm CAV}$ to unit length. For each evaluation image, project $h+5\hat d_{\rm CAV}$ with $W$. Normalize the resulting embedding. The corresponding table rows use the evaluation subset's baseline.

\paragraph{Mahalanobis reweighting.}
For each image and caption embedding, rotate into the $U$ basis. Divide coordinate $k$ by $\max(\sigma_k,10^{-3})$, as implemented in \texttt{unified\_interventions.py}. Normalize each transformed vector before scoring. Mahalanobis reweighting thus modifies both modalities.

\paragraph{Inverted softmax \citep{smith2017offline}.}
For a reference-caption bank $\mathcal B$, define
\[
s_\beta(I,C)=\frac{\exp(\beta I^\top C)}{\sum_{C'\in\mathcal B}\exp(\beta I^\top C')}.
\]
Reference captions come from Winoground and COCO validation captions. We report the sweep $\beta\in\{1,5,10,20,50\}$. The supplied audit script accepts an explicit bank and outputs every setting. The image-specific denominator can be precomputed for a fixed bank.

\paragraph{Random subspace control.}
The supplied intervention script draws a Gaussian matrix with seed $42$ and obtains an orthonormal basis by QR decomposition. Its first $102$ columns define an orthogonal projector $P_{\rm rand}$. The script transforms each projected image as $z'=z+\alpha P_{\rm rand}z$, using $\alpha=5$. It normalizes the result before scoring.

\paragraph{Random pre-projection vector control.}
Draw $r\sim\mathcal N(0,\mathrm{Id}_p)$ and normalize it to unit length. Here $\mathrm{Id}_p$ is the $p\times p$ identity matrix. Project $h+5r$ with $W$ and normalize the resulting embedding. The supplied script uses one seeded direction on the same evaluation subset as the CAV.

\paragraph{Bootstrap summaries.}
The intervention bootstrap uses seed $42$ with the paired-resampling definition in Section~\ref{sec:wino}.

\section{\texorpdfstring{Top-$K$}{Top-K} Approximation}
\label{app:topk}

\paragraph{Audit gallery.}
The default Winoground gallery in \texttt{topk\_audit.py} contains $800$ caption entries from $400$ evaluated pairs. It retains duplicate entries and includes both captions of each pair. Equal-score retrieval ties are broken by gallery order using a stable sort. The script can append a separately encoded external bank. It also implements an annotated-foil exclusion as a diagnostic stress test.

The sweep in Table~\ref{tab:topk} shows that performance worsens as $K$ increases across the tested values. The implementation records Text, Image, and Group scores for each gallery condition so that a gallery test can detect changes to caption rankings.

\section{Learned Bilinear Scorer}
\label{app:bilinear}

The transfer runs in Section~\ref{sec:learned} train on ColorSwap with weight decay in $\{10^{-1},10^{-2},10^{-3},10^{-4},10^{-5}\}$ and evaluate zero-shot on Winoground.

The accompanying \texttt{bilinear\_audit.py} also implements an in-domain protocol. It splits input pairs $60/20/20$ into training, validation, and test sets using seed $26399$. Adam trains $M$ for $100$ epochs with learning rate $10^{-3}$. The loss is the mean softplus of the negatives of the four Text and Image margins, plus $\lambda d^{-2}\|M-\mathrm{Id}_d\|_F^2$, with $\lambda\in\{10^{-1},10^{-2},10^{-3},10^{-4},10^{-5}\}$. Validation Group Score selects $\lambda$, and the first configuration wins ties. The selected matrix is evaluated on the held-out test set.

\section{LLaVA Evaluation Protocol}
\label{app:llava}

LLaVA-1.5-7B~\citep{liu2024improved} uses dense features from a CLIP ViT-L/14 visual encoder. A two-layer MLP connector maps these features to Vicuna-7B inputs.

For each image--caption combination, \texttt{llava\_likelihood.py} constructs the prompt \texttt{Does this image match the following description? \{caption\} Answer yes or no.} inside the checkpoint's conversation format. It sums token log-probabilities for the full continuations \texttt{Yes} and \texttt{No}. Let these continuation log-likelihoods be $\ell_Y$ and $\ell_N$. The score is
\[
s_Y=\ell_Y-\log(\exp\ell_Y+\exp\ell_N).
\]
Each call receives one image and one caption. The four likelihood scores are compared using the strict Text, Image, and Group criteria in Section~\ref{sec:setup}.

\section{Probe Protocol and Caption-Length Heuristic}
\label{app:probe}

\texttt{corrected\_probe\_eval.py} uses the first $3000$ image--caption pairs from the SugarCrepe mirror, subject to dataset size. It concatenates each pooled image embedding with a normalized caption embedding. An $80/20$ split with seed $42$ assigns both captions of each image to the same partition. The MLP has hidden layers $(512,256)$, rectified linear unit activations, Adam optimization, a maximum of $1000$ iterations, and early stopping. Test accuracy averages binary match and mismatch predictions. The length heuristic prefers the shorter caption.

\begin{table}[tbp]
\centering
\caption{SugarCrepe caption-length heuristic and held-out probe accuracy (\%) for CLIP ViT-B/32. $n$ is the category size, and Ties counts caption pairs of equal length.}
\label{tab:probe}
\begin{tabular}{lcccc}
\toprule
\textbf{Category} & $\mathbf{n}$ & \textbf{Length heuristic} & \textbf{Held-out probe} & \textbf{Ties} \\
\midrule
\texttt{add\_obj}     & 1000 & 98.8 & 80.5 & 20 \\
\texttt{add\_att}     &  692 & 99.1 & 70.5 & 8 \\
\texttt{replace\_rel} & 1000 & 54.2 & 61.0 & 505 \\
\texttt{replace\_obj} & 1000 & 55.0 & 58.2 & 775 \\
\texttt{swap\_att}    &  666 & 51.1 & 51.5 & 569 \\
\texttt{replace\_att} &  788 & 51.0 & 50.0 & 660 \\
\texttt{swap\_obj}    &  245 & 52.4 & 50.0 & 221 \\
\bottomrule
\end{tabular}
\end{table}

\section{Qualitative Examples: Successful Recovery}
\begin{figure}[tbp]
    \centering
    \includegraphics[width=\textwidth]{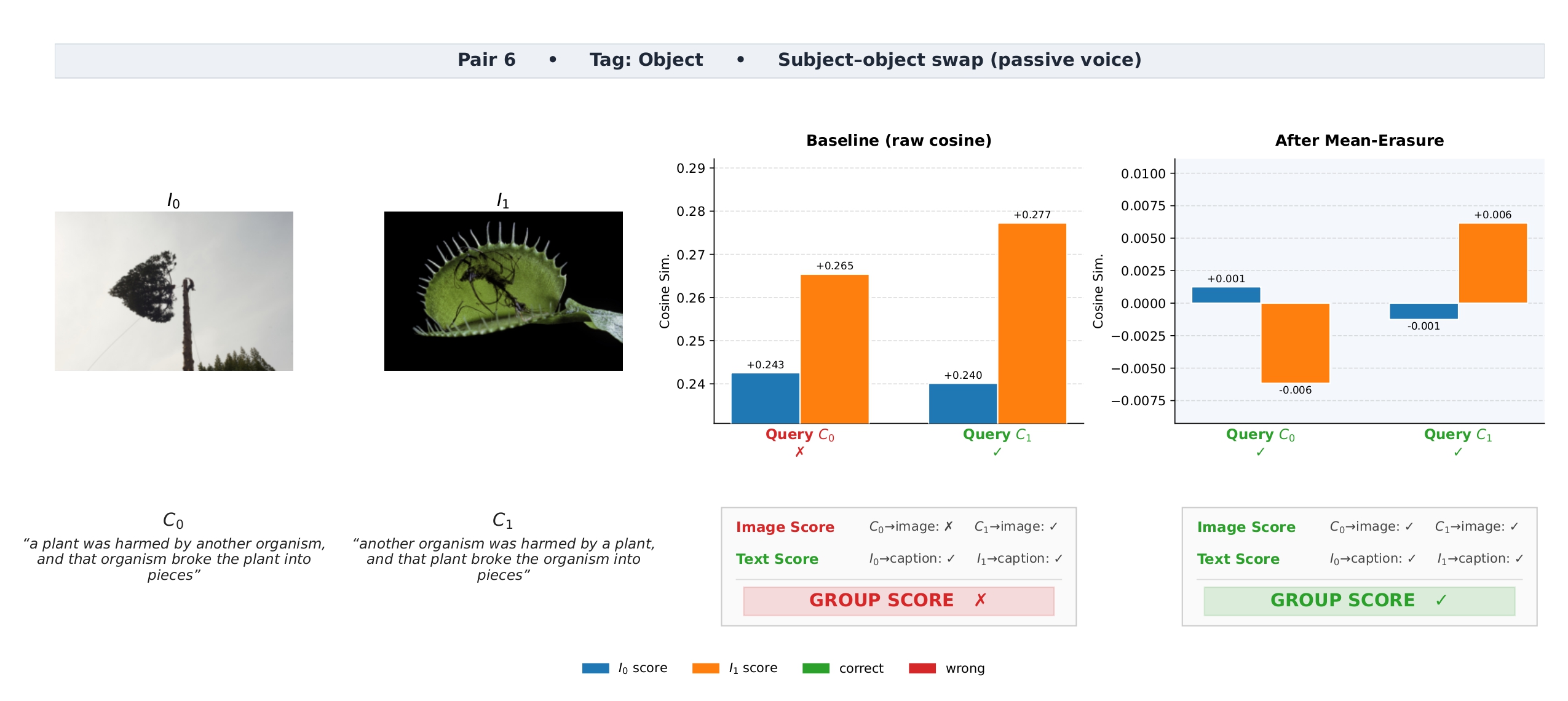}
    \caption{\textbf{Passive-voice Object swap (Pair 6).} ``A plant was
    harmed by another organism'' and\ ``another organism was harmed by a
    plant.'' The plotted baseline Image criterion fails. After caption-mean erasure, both Image and Text criteria pass.}
    \label{fig:qual_success_006}
\end{figure}

\begin{figure}[tbp]
    \centering
    \includegraphics[width=\textwidth]{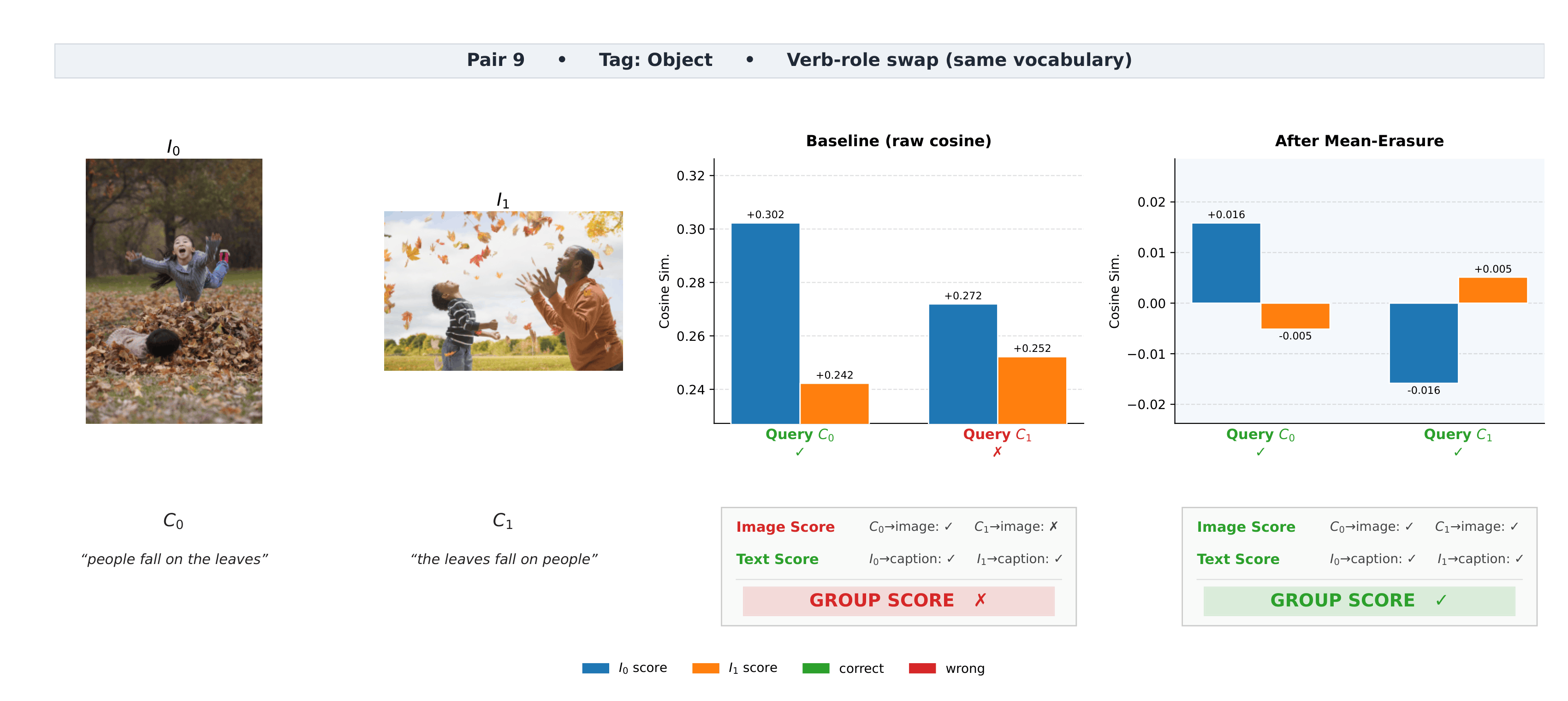}
    \caption{\textbf{Verb-role swap (Pair 9).} ``People fall on the
    leaves'' and\ ``the leaves fall on people'' reuse identical
    vocabulary while swapping the agent of the verb.}
    \label{fig:qual_success_009}
\end{figure}

Figures~\ref{fig:qual_success_006} and~\ref{fig:qual_success_009} add two recovered pairs to Figure~\ref{fig:teaser}. In each, both caption decisions are correct and the baseline Image criterion fails.
\label{app:qual_success}

\section{Qualitative Examples: Residual Group Failures}
\begin{figure}[tbp]
    \centering
    \includegraphics[width=\textwidth]{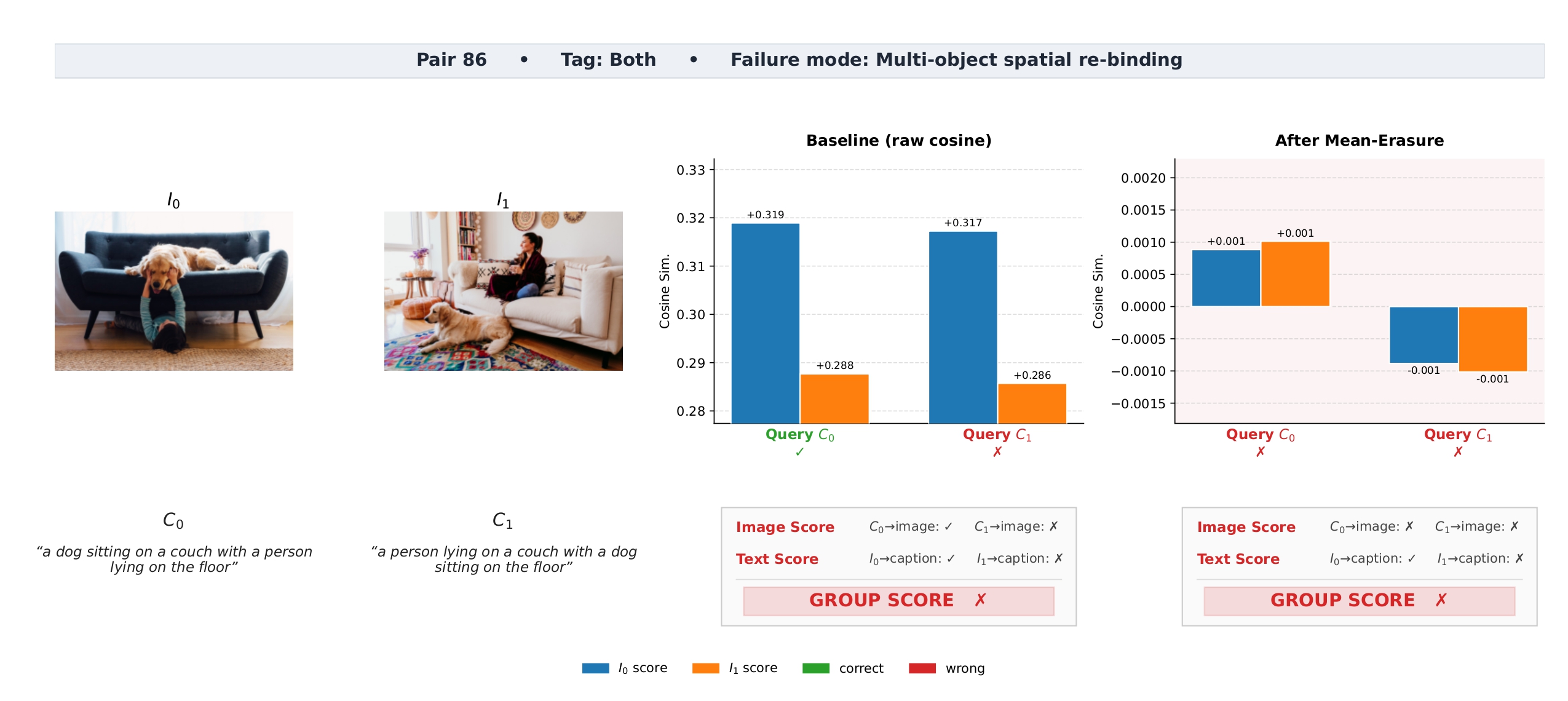}
    \caption{\textbf{Multi-object spatial re-binding (Pair 86, Tag:
    Both).} ``A dog sitting on a couch with a person lying on the
    floor'' and\ the swapped assignment. After erasure both image
    residuals project onto the same side of $\Delta$ with similar
    magnitude. Group Score still fails.}
    \label{fig:qual_fail_086}
\end{figure}

Figure~\ref{fig:qual_fail_086} adds a third unrecovered pair to Figure~\ref{fig:qual_fail}.
\label{app:qual_failures}

\section{Score Offsets and Failure Counts}
\label{app:score_audit}
For $s_{ij}=I_i^\top C_j$, define
\begin{align*}
a_i&=(s_{i0}+s_{i1})/2=I_i^\top\Cmean, &
b_i&=(s_{i0}-s_{i1})/2,\\
m&=b_0-b_1, & o&=|a_0-a_1|.
\end{align*}
The correct-minus-swapped assignment margin is $2m=s_{00}+s_{11}-s_{01}-s_{10}$. The baseline pass criteria and unnormalized erased scores are
\begin{gather*}
\text{Text pass}\iff b_0>0>b_1,\qquad
\text{Image pass}\iff m>o,\\
s'_{ij}=s_{ij}-a_i.
\end{gather*}
Row-centered Image passes when $m>0$. With residual norms $r_i=\|I'_i\|$, normalized mean-erased Image passes when $b_0/r_0>b_1/r_1$.

Here $T,V,G$ denote baseline Text, Image, and Group percentages. $J$ denotes row-centered Image (two-way joint matching), and $V_{\rm ME}$ denotes normalized mean-erased Image. The offset-failure share $F$ is the percentage of baseline Image failures with $0<m\le o$. Table~\ref{tab:audit_wino} shows why Image outcomes under row centering and normalized projection must be distinguished.

For CLIP ViT-B/32, $213/355=60.0\%$ of Image failures have $0<m\le o$. Its $258/400$ joint matches give a $95\%$ Wilson interval
$[59.69,69.03]\%$. The exact one-sided binomial probability is
$3.53\times10^{-9}$ under a $1/2$ null. The $88/400$ additional Group passes have Wilson interval $[18.22,26.32]$ percentage points. The counts are reconstructed by \texttt{audit\_results.py} from the saved aggregate.

\section{Score Nulls and Counterexamples}
\label{app:chance}
\paragraph{Independent-score null.}
For independent and identically distributed continuous scores, the two within-row comparisons involve disjoint pairs of random variables, so Text passes with probability $1/4$. The same argument applies to raw Image Score. Group passes when the two diagonal scores are the largest two of the four, giving probability $1/\binom42=1/6$. Row centering preserves Text, so centered Group equals Text. Centered Image
compares two exchangeable sums. Thus centered $(T,V,G)$ chance rates are $(1/4,1/2,1/4)$ when assignment-sum ties have probability zero.

\paragraph{Text and joint matching.}
For $S=\left(\begin{smallmatrix}0.4&0.2\\0.3&0.2\end{smallmatrix}\right)$, both images prefer caption 0, so Text fails. Yet the diagonal sum $0.6$ exceeds the off-diagonal sum $0.5$, so joint matching passes.

\paragraph{Residual normalization.}
Let $C_0=(\sqrt3/2,1/2,0)$, $C_1=(\sqrt3/2,-1/2,0)$,
$I_0=(0,0.4,\sqrt{0.84})$ and $I_1=(\sqrt{0.96},0.2,0)$. All four vectors have unit length. The caption mean lies along the first axis. The resulting values are
$b_0=0.2$, $b_1=0.1$, $r_0=1$, and $r_1=0.2$. Row-centered Image passes ($0.2>0.1$). Normalized mean-erased Image
fails ($0.2<0.5$). Text and Group fail under both normalization policies.

\paragraph{Partial erasure.}
With $I_i^{(\alpha)}=I_i-\alpha(I_i^\top\Chat)\Chat$, the unnormalized scores are $(1-\alpha)a_i\pm b_i$. For $0\le\alpha\le1$, Image passes iff $m>(1-\alpha)o$. Hence unnormalized Image and Group pass sets grow monotonically.
Text is invariant. This monotonicity result applies to the unnormalized scores. Residual normalization introduces image-dependent row scales.

\paragraph{A symmetric score decomposition.}
Writing $u_i=(-1)^i$ and $v_j=(-1)^j$, every score matrix has the unique form
\[
s_{ij}=\mu+u_i r+v_j c+u_i v_j h,
\]
where $\mu=(s_{00}+s_{01}+s_{10}+s_{11})/4$, $r=(s_{00}+s_{01}-s_{10}-s_{11})/4$,
$c=(s_{00}-s_{01}+s_{10}-s_{11})/4$, and $h=(s_{00}-s_{01}-s_{10}+s_{11})/4$.
Here $T,V,G,J$ denote the Text, Image, Group, and joint-matching pass events. Then $T\iff h>|c|$, $V\iff h>|r|$, $G\iff h>\max(|r|,|c|)$, and $J\iff h>0$. Row centering removes $\mu,r$. Column centering removes $\mu,c$.
Double centering retains only the interaction term $u_i v_jh$.
Column centering gives $G_{\rm col}\iff V$, and double centering gives $G_{\rm double}\iff J$.

\section{Multi-Candidate Rankings and Assignment Sums}
\label{app:larger}
Subtracting one constant per row preserves within-row rankings for any matrix size. The identity between centered Group and raw Text is specific to two columns and two rows. For example,
\[
S=\begin{pmatrix}2&1&0\\99&100&-100\\0&0&1\end{pmatrix}
\]
has each diagonal entry as its row maximum. After row centering, the first column is $(1,66,-1/3)^\top$, so the first caption retrieves the second image. Group therefore fails despite Text passing.

For arbitrary numbers of candidates, joint assignment sums are invariant to additive row and column offsets: every bijection contains each row and column once, so all assignments receive the same added constant.

\section{Evaluation Protocol and Evidence Coverage}
\label{app:replication_notes}
The five identifiers used by the retained Winoground script are:
\begin{center}\small
\begin{tabular}{ll}
\toprule
Model & Checkpoint identifier\\
\midrule
CLIP ViT-B/32 (OpenAI) & \texttt{openai/clip-vit-base-patch32}\\
CLIP ViT-B/32 (LAION) & \texttt{laion/CLIP-ViT-B-32-laion2B-s34B-b79K}\\
CLIP ViT-L/14 (OpenAI) & \texttt{openai/clip-vit-large-patch14}\\
SigLIP-base & \texttt{google/siglip-base-patch16-224}\\
SigLIP-SO400M & \texttt{google/siglip-so400m-patch14-384}\\
\bottomrule
\end{tabular}
\end{center}
\paragraph{Extraction protocol.}
The dataset mirror is \path{haideraltahan/wds_winoground}. The documented Winoground protocol uses RGB images, the checkpoint processor, float32 feature extraction, and normalized pooled image and text embeddings. SigLIP text inputs use fixed-length padding. The supplied extraction code can save checkpoint and dataset metadata with scores and embeddings.

\paragraph{Evidence coverage.}
The five-checkpoint Winoground scores and the unrelated-direction control come from the saved aggregate \path{results/winoground_existing_run.json}. The aggregate audit reconstructs integer counts at $n=400$, computes Wilson intervals for recovered pairs, and checks the score identities on synthetic unit vectors.

The ColorSwap, swap-tag, partial-strength, noun-prompt, fixed-correction, bilinear, LLaVA, probe, and top-$K$ results come from earlier evaluation runs and are reported as aggregate scores. Their uncertainty estimates are the bootstrap tails and the ColorSwap Wilson intervals, which use counts reconstructed from rounded scores. The random-subspace script uses $\alpha=5$. The accompanying extraction and scoring code implements each evaluation. The main text's cost claim counts vector operations.

\paragraph{Noun-prompt control.}
The noun-prompt control tokenizes caption~0 with the Natural Language Toolkit and keeps the tags that begin with \texttt{NN}. It joins the distinct nouns with ``and'' and forms the prompt \texttt{A photo of a \{nouns\}}; an empty noun set becomes ``objects''. CLIP's text tower and projection encode the prompt, and the normalized vector is erased from both images.

\paragraph{Aggregate audit.}
\path{code/audit_results.py} reconstructs the five-checkpoint counts and uncertainty intervals. The consistency audit (\path{code/revision_audit.py}) also verifies that every main-table row matches the saved aggregate and checks the paired projection's invariances. Both audits run on included results or synthetic unit vectors. The ancillary files include their generated outputs and exact commands.

\paragraph{Ties and numerical stability.}
The supplied audit implementation in \texttt{metrics.py} converts vectors and score arithmetic to float64. Input normalization raises an error for any embedding norm below $10^{-12}$. For an erasure direction with norm at most $10^{-12}$, it uses a zero direction, leaving the image unchanged. For a projection residual with norm at most $10^{-12}$, it returns a zero scoring vector. The associated score row is then zero, which extends the implementation beyond the nonzero-vector domain of normalized cosine.

The pass function uses strict differences greater than zero. Exact ties fail. For erasure along caption $C_j$, the projection gives $I_i^{\prime\top}C_j=0$. The audit sets this score column to zero before comparisons. Its strict ties give $\mathrm{Image}=\mathrm{Group}=0$. Each reported percentage uses all input benchmark pairs as its denominator. The supplied paired audit records the count of residual norms at or below $10^{-12}$. It computes sensitivity at comparison tolerances $\tau\in\{0,10^{-10},10^{-8},10^{-6}\}$. At tolerance $\tau$, a comparison passes only when its difference exceeds $\tau$.

\section{Notation}
\label{app:notation}
\begin{center}
\begin{tabular}{lp{0.6\textwidth}}
\toprule
\textbf{Symbol} & \textbf{Meaning} \\
\midrule
$I_i,C_j$ & Normalized image and caption embeddings in $\mathbb R^d$ \\
$\Cmean,\Delta$ & Half-sum and half-difference of the candidate captions \\
$I'_i,r_i$ & Projected image and its residual norm $\|I'_i\|$ \\
$I_{\rm mean}$ & Mean of the two image embeddings (Equation~\ref{eq:image_erasure}) \\
$T,V,G,J$ & Text, Image, Group, and joint-matching (row-centered Image) pass events \\
$s_{ij}$ & Baseline cosine score $I_i^\top C_j$ \\
$a_i,b_i$ & Common score component and signed caption contrast for image $i$ \\
$m,o$ & Joint-matching half-margin $b_0-b_1$ and offset $|a_0-a_1|$ \\
$\alpha,K$ & Erasure strength and number of retrieved captions \\
\bottomrule
\end{tabular}
\end{center}

\end{document}